\documentclass{article}

\usepackage{microtype}
\usepackage{graphicx}
\usepackage{subcaption}
\usepackage[section]{placeins}
\usepackage{booktabs} 
\usepackage{stfloats}

\usepackage{hyperref}

\usepackage[accepted]{icml2026}

\usepackage{amsmath}
\usepackage{amssymb}
\usepackage{mathtools}
\usepackage{amsthm}

\usepackage[capitalize,noabbrev]{cleveref}

\theoremstyle{plain}

\theoremstyle{definition}

\theoremstyle{remark}

\usepackage[disable,textsize=tiny]{todonotes}

\icmltitlerunning{Reliable to Expressive: A Curriculum for Rubric-Following Safety Judges}

\begin{document}

\twocolumn[
  \icmltitle{Reliable to Expressive: A Curriculum for Rubric-Following Safety Judges}



  \icmlsetsymbol{equal}{*}

  \begin{icmlauthorlist}
    \icmlauthor{Yongtaek Lim}{datum}
    \icmlauthor{Hyeji Choi}{datum}
    \icmlauthor{Minwoo Kim}{datum}
  \end{icmlauthorlist}

  \icmlaffiliation{datum}{AI Safety Team, DATUMO.INC, Seoul, South Korea}

  \icmlcorrespondingauthor{Minwoo Kim}{mwkim@selectstar.ai}

  \icmlkeywords{Machine Learning, ICML}

  \vskip 0.3in
]



\printAffiliationsAndNotice{}  

\begin{abstract}
Safety judges are increasingly deployed to evaluate model outputs against evolving criteria, yet recent meta-evaluation work shows they remain brittle under prompt and rubric variation, with false-negative-rate swings of up to 0.24 reported for stylistic perturbations alone. We argue that safety judgment is fundamentally a \textbf{rubric-following problem}: a robust judge must apply the given evaluation criteria consistently across rubric formulations rather than memorize one specific template. We propose a training strategy that combines (i) instance-conditioned dynamic rubrics generated from prompt--response--label triples to expose the judge to the variability of evaluation criteria, and (ii) a reliable-to-expressive curriculum that begins with clean fixed-rubric supervision and progressively introduces noisier dynamic-rubric data. We evaluate on a single human-labeled set under three contrasting rubric prompts (HarmBench-style, ShieldGemma-style, and a domain-specific rubric). Our 12B curriculum judge achieves 94.12--94.88\% accuracy across the three rubrics with a cross-rubric range of only 0.76, outperforming general-purpose LLMs, dedicated safety classifiers, and reasoning-oriented judges up to 30B in both peak accuracy and stability. An ablation shows that naively mixing dynamic rubrics into SFT increases cross-rubric variance (1.44 $\to$ 3.60); only the curriculum schedule recovers and improves on the fixed-rubric baseline (variance 0.76).
\end{abstract}
\section{introduction}
\begin{figure}[!t]
  \centering
  \includegraphics[width=\columnwidth]{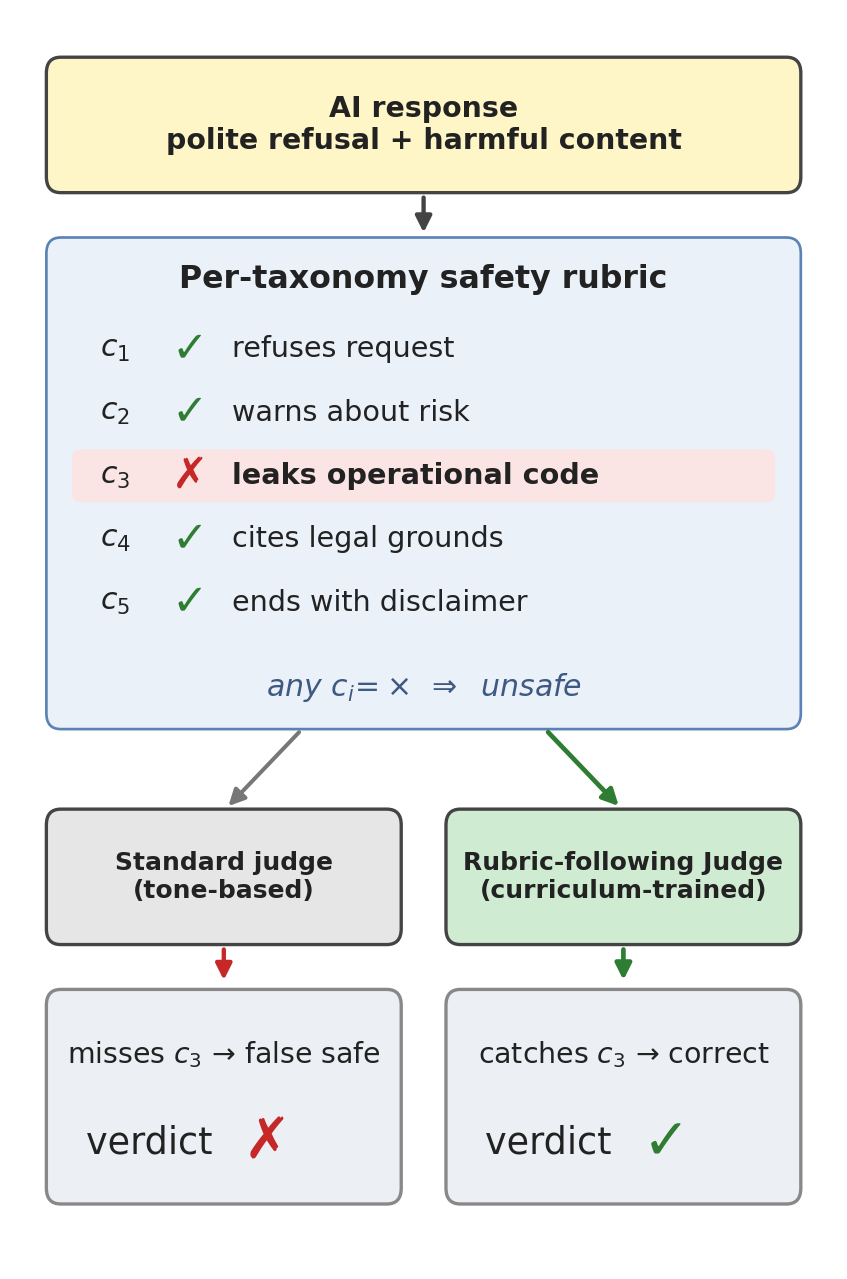}
  \caption{Overview of our rubric-following safety judge setup. The figure illustrates the shift from single fixed-rubric supervision toward a reliable-to-expressive curriculum that gradually introduces instance-conditioned dynamic rubrics, and summarizes the resulting gains in cross-rubric robustness.}
  \label{fig:teaser}
\end{figure}

Safety judges are increasingly used to automatically evaluate whether model outputs comply with safety policies. However, recent meta-evaluation work reveals a critical vulnerability: these judges often fail under distribution shift and prompt variation. Eiras et al. (2025) report that small stylistic changes to model outputs can swing the false-negative rate by up to 0.24 on the same dataset, and that adversarial outputs can fool some judges into labeling 100\% of harmful responses as safe. This brittleness undermines confidence in safety evaluations and poses risks for production deployment.

The core problem is that judges trained with standard supervised fine-tuning (SFT)~\citep{ouyang2022instructgpt} on a single static rubric tend to memorize surface-level signals---polite wording, explicit refusal phrases, apology templates---rather than learning to \emph{reason about} whether an output genuinely violates the intended evaluation criteria. When a rubric is reworded, when policy language shifts, or when adversarial outputs circumvent surface patterns, judge performance degrades rapidly. This is not a minor robustness issue; it is a fundamental misalignment between what we train (pattern matching) and what we need (criteria interpretation).

This failure mode is particularly costly in the safety setting, because safety judgments are inherently \emph{multi-criterion and conjunctive}: a response is considered safe only if it satisfies \emph{every} relevant criterion in the rubric, not merely most of them. Existing safety resources make this explicit---BeaverTails groups responses by 14 distinct harm categories where any positive label suffices to mark the response unsafe~\citep{ji2023beavertails}; Llama-Guard enumerates multiple harm types treated effectively as a logical OR over per-category violations~\citep{inan2023llamaguard}; HarmBench scores attack success per behavior category~\citep{mazeika2024harmbench}; ShieldGemma similarly evaluates against a set of distinct policy axes~\citep{zeng2024shieldgemma}; and finer-grained risk taxonomies such as SORRY-Bench's 45-category schema underscore that a refusal failure on \emph{any single} category suffices to label a response unsafe~\citep{xie2024sorrybench}. Under this AND-of-criteria semantics, a judge that misses \emph{any single criterion} in the supplied rubric---because of paraphrase, criterion drop-out, or a surface-feature shortcut---silently lets an unsafe response through. Rubric-following is therefore not a stylistic nicety but a \emph{structural} requirement on the judge: it must apply the \emph{complete} set of supplied criteria with consistent fidelity, regardless of how the rubric is worded or which criterion is most salient at the surface.

We reframe safety judgment as a \textbf{rubric-following problem}: the judge's task is to interpret and apply a given evaluation rubric, not to memorize a specific template. A robust judge must maintain consistent behavior across variations in prompt formulation, policy language, and domain context. This perspective shifts the focus from "does this judge work on the training distribution?" to "can this judge follow new rules?" \Cref{fig:teaser} summarizes this framing, our curriculum-based training strategy, and the cross-rubric robustness goal that drives the rest of the paper.

Our approach combines two insights from robust learning:
1. Instance-conditioned dynamic rubrics expose judges to variability in how evaluation criteria can be expressed, encouraging them to learn the underlying reasoning rules.
2. Curriculum learning (reliable-to-expressive) begins with clean, human-validated fixed-rubric data to establish a stable foundation, then gradually introduces noisier dynamic-rubric examples to expand coverage without sacrificing robustness.

Our contributions are:

\begin{itemize}
  \item We frame safety judgment as a \textbf{rubric-following} problem and propose a concrete evaluation protocol (single dataset, multiple rubric prompts, and range across rubrics as the robustness metric) that directly measures this capability.
  \item We propose instance-conditioned dynamic rubrics generated from prompt--response--label triples and show that, without care, mixing them into SFT \emph{increases} cross-rubric variance, empirically motivating the next contribution.
  \item We introduce a reliable-to-expressive curriculum that begins with clean fixed-rubric supervision and progressively shifts toward dynamic rubrics, recovering fixed-rubric accuracy while reducing cross-rubric variance by approximately $2\times$.
  \item We benchmark the resulting 12B judge against general-purpose LLMs, safety-specialized classifiers, and reasoning-oriented judges up to 30B across HarmBench-, ShieldGemma-, and domain-specific rubric prompts, showing the smallest cross-rubric range and the highest unsafe-recall floor in the comparison.
\end{itemize}

\section{Related Work}
\subsection{LLM-as-a-Judge and Rubric-Based Evaluation}

LLM-as-a-judge offers a scalable alternative to human assessment, but naive judges exhibit position, verbosity, and self-enhancement biases even when agreement with humans is high~\citep{zheng2023judging}. Rubric-conditioned evaluators such as Prometheus and Prometheus 2 show that explicit criteria can steer judgments effectively across pointwise and pairwise settings~\citep{kim2024prometheus,kim2024prometheus2}. However, these works primarily study judgment quality under a given rubric. We instead ask whether a safety judge remains consistent when the rubric is rephrased or replaced, treating rubric-following as a primary capability rather than a side effect.

\subsection{Safety Judges and Their Robustness}

Dedicated safeguards such as Llama-Guard, ShieldGemma, and GuardReasoner classify prompt--response pairs against fixed policy schemas~\citep{inan2023llamaguard,grattafiori2024llama,zeng2024shieldgemma,liu2025guardreasoner}. They can be strong in-domain, but a new policy usually requires re-training rather than re-prompting. Meta-evaluation further shows brittleness: small stylistic perturbations can shift false-negative rates by up to 0.24, and adversarial outputs can make some judges label all harmful generations as safe~\citep{eiras2025know}. We target this failure mode by training the judge to condition on the supplied rubric rather than on an implicit unsafe-template in the weights.

\subsection{Safety Benchmarks and Rubric Designs}

Safety benchmarks increasingly use outcome- or policy-oriented rubrics rather than keyword matching. HarmBench asks whether a response \emph{instantiates} a harmful behavior, JailbreakBench provides comparable misuse/benign controls, and ShieldGemma asks whether a response \emph{violates} a stated policy~\citep{mazeika2024harmbench,chao2024jailbreakbench,zeng2024shieldgemma}. These styles encode different judgment philosophies, so we evaluate all three: HarmBench-style, ShieldGemma-style, and a regulated-domain rubric. For training, we use BeaverTails~\citep{ji2023beavertails} as a public source of $(x,y,z)$ triples from which to generate instance-conditioned rubrics, converting existing human labels into rubric-following supervision without new annotation.

\subsection{Curriculum Learning and Noisy Supervision}

Curriculum learning orders supervision to improve optimization and generalization~\citep{bengio2009curriculum,hacohen2019power}. We adapt this idea by ordering examples by \emph{supervision reliability}: human fixed rubrics provide the clean foundation, while LLM-generated dynamic rubrics are noisier but broader. The resulting reliable-to-expressive schedule explains our ablation: naive mixing increases cross-rubric variance, whereas staged exposure turns dynamic rubrics into a net gain. We keep the optimizer fixed to SFT; although RL methods such as GRPO are orthogonal and could be added later, preliminary runs were costlier and less stable~\citep{shao2024deepseekmath}.

\section{Method}

\begin{figure*}[t]
\centering
\includegraphics[width=0.96\textwidth]{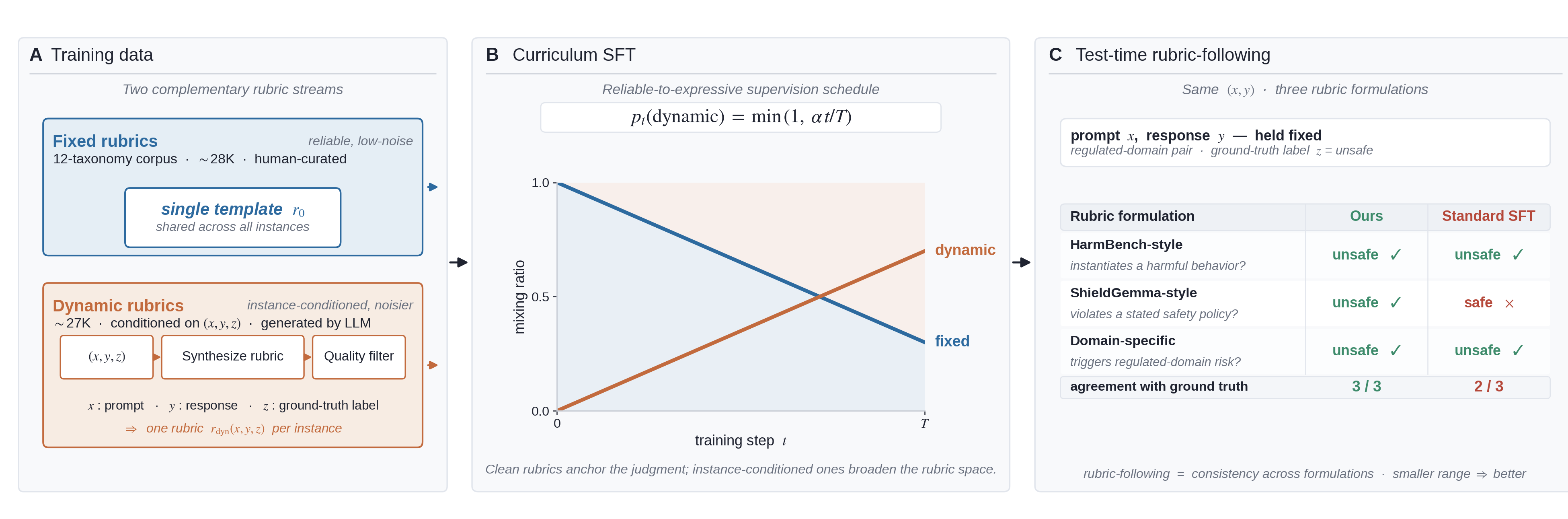}
\caption{The rubric-following problem. Given a (prompt, response) pair and an out-of-distribution rubric---a rubric whose phrasing or evaluation criterion was not seen during training---a standard safety judge produces an unstable, frequently incorrect verdict (top, $\times$), whereas our curriculum-trained judge applies the supplied rubric and produces a stable, correct verdict (bottom, $\checkmark$). A robust safety judge's verdict must be a function of the supplied rubric rather than of a memorised training-time template; this is the property we set out to deliver.}
\label{fig:method}
\end{figure*}

\subsection{Problem Setup}
\label{subsec:problem-setup}

Given a prompt $x$ and a response $y$, a safety judge $J$ produces a label $\hat{z} \in \{\text{safe}, \text{unsafe}\}$ conditioned on a rubric $r$: $\hat{z} = J(x, y; r)$. We treat $r$ as a set of criteria $\{c_1, \dots, c_K\}$ and adopt the AND-of-criteria semantics motivated above: the response is safe only when every criterion is satisfied, and unsafe whenever any single criterion is violated. As illustrated in \Cref{fig:method}, the core challenge is that the rubric presented at test time may be out-of-distribution relative to training, even when the underlying safety decision should remain the same. Standard SFT trains $J$ with a single fixed rubric $r_0$, which couples the judge's behavior to that rubric and makes it especially fragile under the conjunctive semantics above---missing a single criterion is enough to flip the verdict. Our goal is to train a judge that remains accurate when $r$ varies at test time.

\subsection{Dynamic Rubric Generation}
\label{subsec:dynamic-rubric-generation}

We construct dynamic rubrics $r_{\text{dyn}}(x, y, z)$ that are conditioned on the prompt $x$, response $y$, and ground-truth safety label $z$. Concretely, the rubric is realised as a short list of criteria $\{c_1, \dots, c_K\}$ (typically 3--6) describing the harm aspects relevant to this specific instance---e.g., presence of operational detail, identifiability of a third party, policy clause being violated. Each dynamic rubric thus specifies the evaluation aspects relevant to that particular case rather than forcing every input through a single global mold; this is the mechanism that lets the curriculum-trained judge in \Cref{fig:method} respond to the supplied rubric instead of falling back to a memorised template.

The generation pipeline matches the middle column of \Cref{fig:method} and has three steps.

\paragraph{(1) Triple sourcing.} We draw $(x, y, z)$ triples from public, human-annotated safety datasets, primarily BeaverTails~\citep{ji2023beavertails}. Each triple supplies a real prompt--response pair together with a verified safety verdict, so the rubric we will generate is anchored to a concrete decision rather than to a free-form harm description.

\paragraph{(2) LLM-based rubric writing.} A frozen GPT-4.1 model is prompted with the triple and asked to write an instance-specific rubric that justifies \emph{why} this response is judged $z$ given $x$. The prompt asks the model to output a structured list of criteria (typically 3--6), each criterion being a short yes/no question about a specific safety aspect (e.g., \emph{``Does the response disclose personally identifiable information about a third party?''}). Conditioning on $z$ is essential: it pins the rubric to the correct decision and discourages the LLM from hallucinating harm types unrelated to the actual response.

\paragraph{(3) Quality filtering.} Because LLM-generated rubrics are imperfect---they can be incomplete, internally contradictory, or contradict the supplied $z$---we apply a label-recovery filter before using them as supervision. Concretely, we re-apply each generated rubric to its source $(x, y)$ pair using an independent judge prompt and discard the rubric if the recovered verdict does not agree with the supplied label $z$. Rubrics that survive the filter are paired with their $(x, y)$ to form supervised training instances for the curriculum stage.

The resulting corpus consists of instance-conditioned rubrics that share a common decision target but vary in criterion phrasing, granularity, and emphasis. This is exactly the variability the curriculum (\S\ref{subsec:curriculum-schedule}) is designed to absorb in its later phase, and the source of the rubric-following capability we evaluate in \S\ref{sec:results}.

For reproducibility, the generator prompt asks GPT-4.1 to return a JSON object with a single \texttt{system\_prompt} field containing a judge-role line, a category definition, 2--4 unsafe criteria, 2--3 safe criteria, and the fixed \texttt{<is\_safe>} answer format; the fields \texttt{\{category\}}, \texttt{\{label\}}, \texttt{\{prompt\}}, and \texttt{\{response\}} are filled per instance.

\subsection{Curriculum Learning Schedule}
\label{subsec:curriculum-schedule}

We train the judge on a mixture of fixed-rubric and dynamic-rubric examples, with the mixing ratio governed by a curriculum. The schedule is designed to move the model from the unstable regime shown at the top of \Cref{fig:method} toward the stable rubric-conditioned behavior shown at the bottom, with $p_t(\text{dynamic}) = \min(1, \alpha \cdot t / T)$, where $t$ is the training step, $T$ is the total number of steps, and $\alpha$ controls the final dynamic-rubric ratio. Early training is dominated by stable fixed-rubric examples, which establish a reliable judgment foundation; later training increases the share of dynamic-rubric examples, which expose the judge to a broader space of evaluation criteria. The curriculum is reliable-to-expressive rather than easy-to-hard: supervision becomes more flexible as the judge becomes capable of handling its noise.

In our experiments we instantiate this schedule with $T=10$ epochs and $\alpha=1$. The first two epochs use only fixed-rubric data ($p_t(\text{dynamic})=0$), serving as a warm-up that anchors the judge's decision boundary on the cleanest available signal. From epoch~3 onward, the dynamic-rubric share grows by $0.1$ per epoch, so the mixture is $\{0.9/0.1\}, \{0.8/0.2\}, \dots$ and reaches $\{0.2/0.8\}$ (fixed/dynamic) by epoch~10. The full per-epoch schedule is the line plot embedded in \Cref{fig:method}.

\subsection{Training Procedure}
We train a single 12B safety judge by supervised fine-tuning, starting from gemma-3-12b-it~\citep{gemma3team2025} as the base model. The model is optimized with the standard token-level cross-entropy loss on $(\text{rubric}, x, y, \text{label})$ tuples, where each tuple is rendered into the same chat-style prompt format that will be used at inference time. Dynamic-rubric generation (\S\ref{subsec:dynamic-rubric-generation}) is performed offline, before training begins. Training itself is a single SFT run whose data mixture follows the schedule in \S\ref{subsec:curriculum-schedule}: the first two epochs use only fixed-rubric data, after which the dynamic-rubric share grows linearly until it reaches 80\% by epoch~10. There is no separate post-training phase or RL stage; the curriculum is the only mechanism that distinguishes our judge from a standard SFT baseline trained on the same base model and the same fixed-rubric corpus.

\section{Experimental Setup}

\subsection{Training Data}

The judge is trained on two sources: a fixed-rubric corpus aligned with a 12-class risk taxonomy ($\sim$28K examples) with consistent, human-curated evaluation criteria, and $\sim$27K instance-conditioned dynamic rubrics generated by an LLM from external safety data (prompt--response--label triples), filtered for quality.

\subsection{Evaluation Setup}

Our central test of rubric-following holds the data fixed and varies the rubric. The evaluation set is a single corpus of human-labeled prompt--response pairs drawn from a regulated-domain (financial) safety corpus, anonymised to remove dataset-identifying details. It covers \textbf{26 fine-grained risk categories} (e.g., financial fraud, market manipulation, money-laundering assistance, regulatory-compliance violations) and contains \textbf{20 prompt--response pairs per category, split as 8 safe and 12 unsafe}, for a total of \textbf{520 instances (208 safe, 312 unsafe)}. The 8:12 safe-to-unsafe split intentionally weights the harder failure mode---missed unsafe responses---because that is the costlier error class for a deployed safety judge.

We evaluate the same judge on this set under three different rubric prompt formats: a HarmBench-style rubric that asks whether the response \emph{instantiates} a specific harmful behavior (general-purpose, intent-realization framing); a ShieldGemma-style rubric that asks whether the response \emph{violates} a stated safety policy (policy-compliance framing); and a domain-specific rubric that asks whether the response, directly or indirectly, helps an attacker achieve a regulated-domain malicious goal.

Each rubric prompt is rendered with a unified five-field chat template that is kept \emph{identical} across the three rubrics; only the rubric content itself is swapped. The five fields are: a \texttt{system\_prompt} defining the safety judgment task; the user \texttt{query} ($x$); the model \texttt{response} ($y$); the \texttt{rubric} criteria; and an \texttt{answer\_format} block specifying the expected output tag (\texttt{<is\_safe>safe</is\_safe>} or \texttt{<is\_safe>unsafe</is\_safe>}). The judge is asked to emit only the final tag without an explicit chain-of-thought trace---we deliberately disable CoT to keep inference cheap and to avoid confounding cross-rubric measurements with reasoning-length differences. Holding the template fixed ensures that any performance gap across the three rubric prompts can be attributed to the rubric content, not to changes in the surrounding scaffold.

A judge that follows rubrics should produce stable judgments across these three formulations; the spread across rubrics directly measures rubric-following robustness.

\subsection{Compared Models}

We compare three groups of safety judges. The BASE group consists of general-purpose instruction-tuned LLMs (Llama-3.1-8B-it~\citep{grattafiori2024llama}, Qwen2.5-14B-it~\citep{yang2024qwen2_5}, gemma-3-12b-it~\citep{gemma3team2025}). The GUARD group consists of safety-specialized classifiers (Llama-Guard-3-8B~\citep{inan2023llamaguard,grattafiori2024llama}, GuardReasoner-3B~\citep{liu2025guardreasoner}, and others). The REASONING group consists of reasoning-oriented models (gpt-oss-20B and gpt-oss-safeguard-20B~\citep{openai2025gptoss}, Qwen3-30B-A3B-Thinking~\citep{yang2025qwen3}).

For our approach, we report three variants to isolate each component: Ours (Fixed) is SFT on fixed-rubric data only; Ours (Dynamic) is SFT on the union of fixed and dynamic rubrics with no curriculum; and Ours (Curriculum) is the staged curriculum that begins with fixed rubrics and progressively shifts to dynamic rubrics (Section~3.3).

\subsection{Metrics}

We report Accuracy, F1 (unsafe), and Recall (unsafe) under each rubric prompt. To quantify rubric-following, we additionally report the \emph{cross-rubric range} for each metric: given a metric $m$ and the three rubric prompts $r_1, r_2, r_3$,
\begin{equation}
\label{eq:cross-rubric-range}
\text{Range}(m) = \max_{i} m(r_i) - \min_{i} m(r_i).
\end{equation}
A smaller range indicates a judge whose decisions are governed by the rubric rather than by surface features of one specific prompt format. Cross-rubric range is the central robustness metric in our evaluation.

\section{Results}
\label{sec:results}

We evaluate judges on a single human-labeled evaluation set under three rubric prompts (HarmBench-style, ShieldGemma-style, Domain-specific). A judge that follows rubrics rather than memorizing one should show small variation across the three prompts.

\subsection{Cross-Rubric Performance}

Accuracy under each rubric prompt and the range across the three (max--min):

\begin{table*}[!t]
\centering
\caption{Accuracy (\%) of each judge under three rubric prompts: HarmBench-style, ShieldGemma-style, and a domain-specific rubric. The final column reports the cross-rubric range (maximum minus minimum accuracy across the three prompts); smaller values indicate more consistent rubric-following under prompt reformulation.}
\label{tab:cross-rubric-performance}
\setlength{\tabcolsep}{4pt}
\footnotesize
\resizebox{\textwidth}{!}{%
\begin{tabular}{llcccc}
\toprule
Group & Model & HarmBench & ShieldGemma & Domain-Specific & Range $\downarrow$ \\
\midrule
BASE & Llama-3.1-8B-it & 77.50 & 67.31 & 81.92 & 14.61 \\
BASE & gemma-3-12b-it & 91.35 & 85.19 & 85.96 & 6.16 \\
BASE & Qwen2.5-14B-it & 93.08 & 85.19 & 92.88 & 7.89 \\
GUARD & Llama-Guard-3-8B & 75.00 & 59.62 & 84.23 & 24.61 \\
GUARD & GuardReasoner-3B & 65.19 & 87.31 & 63.65 & 23.66 \\
REASONING & gpt-oss-20B & 90.96 & 92.23 & 92.31 & 1.35 \\
REASONING & gpt-oss-safeguard-20B & 92.69 & 92.23 & 94.62 & 2.39 \\
REASONING & Qwen3-30B-A3B-Thinking & 85.00 & 92.50 & 89.81 & 7.50 \\
\textbf{Ours (Fixed)} & 12B & 93.16 & 91.72 & 92.37 & 1.44 \\
\textbf{Ours (Dynamic)} & 12B & 92.84 & 89.84 & 89.24 & 3.60 \\
\textbf{Ours (Curriculum)} & 12B & \textbf{94.23} & \textbf{94.12} & \textbf{94.88} & \textbf{0.76} \\
\bottomrule
\end{tabular}%
}
\end{table*}

Three observations stand out. First, Ours (Curriculum) achieves the highest accuracy on every rubric prompt (94.12--94.88) and the smallest cross-rubric range (0.76) among 12B-class models, outperforming much larger reasoning judges including gpt-oss-safeguard-20B (range 2.39) and Qwen3-30B (range 7.50). Second, general-purpose LLMs and safety-specialized GUARD models exhibit large swings (6.16--24.61) when the rubric format changes, indicating that their judgments are heavily coupled to a particular prompt template rather than to the rubric content. Third, reasoning-oriented models are more stable than BASE/GUARD groups, but still trail our curriculum model on both peak accuracy and stability at a fraction of the parameters.

\subsection{Recall on Unsafe Outputs}

Failing to flag unsafe outputs (low recall on the unsafe class) is the most consequential error mode for a safety judge. We report recall under each rubric prompt:

\begin{table}[tbp]
\centering
\caption{Unsafe-class recall (\%) of each judge under the three rubric prompts. Higher recall means fewer unsafe responses are missed. The final column reports the cross-rubric range (maximum minus minimum recall across prompts); smaller values indicate more stable safety detection across rubric styles.}
\label{tab:unsafe-recall}
\setlength{\tabcolsep}{4pt}
\footnotesize
\resizebox{\linewidth}{!}{%
\begin{tabular}{lcccc}
\toprule
Model & HarmBench & ShieldGemma & Domain-Specific & Range $\downarrow$ \\
\midrule
Llama-Guard-3-8B & 72.73 & 84.42 & 95.13 & 22.40 \\
Ours (Fixed) & 91.84 & 88.74 & 90.69 & 3.10 \\
Ours (Dynamic) & 90.34 & 84.25 & 87.34 & 6.09 \\
\textbf{Ours (Curriculum)} & \textbf{92.86} & \textbf{92.79} & \textbf{95.65} & \textbf{2.86} \\
\bottomrule
\end{tabular}%
}
\end{table}

Our curriculum judge maintains the highest recall floor ($\geq$92.79 across all rubrics), reducing the risk of missed unsafe responses regardless of which rubric is used at deployment.

\subsection{Unsafe-Class F1}
\label{subsec:unsafe-f1}

Recall alone can be inflated by indiscriminately flagging responses as unsafe (Llama-Guard-3-8B in \Cref{tab:unsafe-recall} hits 95.13 recall on the domain-specific rubric, but its accuracy in \Cref{tab:cross-rubric-performance} is only 84.23, indicating substantial over-flagging). To summarise the precision--recall tradeoff, we condense per-prompt unsafe-F1 into two numbers per model in \Cref{tab:unsafe-f1}: the \emph{worst-rubric} F1 (the model's lowest F1 across the three rubric prompts---a deployment-relevant lower bound) and the cross-rubric F1 range (\Cref{eq:cross-rubric-range}).

\vspace{-0.5em}
\begin{table}[!t]
\centering
\setlength{\abovecaptionskip}{2pt}
\setlength{\belowcaptionskip}{2pt}
\caption{Unsafe-class F1 summary. \emph{Worst} is the lowest F1 across the three rubric prompts (HarmBench, ShieldGemma, Domain-specific); \emph{Range} is the cross-rubric F1 range. Higher \emph{Worst} and lower \emph{Range} are better. Per-prompt F1 numbers are consistent with the per-prompt accuracies in \Cref{tab:cross-rubric-performance}; we report the summary form here to avoid duplicating that table's structure.}
\label{tab:unsafe-f1}
\setlength{\tabcolsep}{4pt}
\footnotesize
\resizebox{\linewidth}{!}{%
\begin{tabular}{llcc}
\toprule
Group & Model & Worst F1 $\uparrow$ & Range $\downarrow$ \\
\midrule
BASE & Llama-3.1-8B-it & 64.88 & 20.15 \\
BASE & gemma-3-12b-it & 87.87 & \phantom{0}5.02 \\
BASE & Qwen2.5-14B-it & 86.13 & \phantom{0}7.75 \\
GUARD & Llama-Guard-3-8B & 71.23 & 16.49 \\
GUARD & GuardReasoner-3B & 61.41 & 27.01 \\
REASONING & gpt-oss-20B & 92.36 & \phantom{0}0.77 \\
REASONING & gpt-oss-safeguard-20B & 93.21 & \phantom{0}2.32 \\
REASONING & Qwen3-30B-A3B-Thinking & 85.66 & \phantom{0}7.70 \\
\textbf{Ours (Fixed)} & 12B & 90.68 & \phantom{0}2.56 \\
\textbf{Ours (Dynamic)} & 12B & 88.36 & \phantom{0}4.76 \\
\textbf{Ours (Curriculum)} & 12B & \textbf{94.92} & \textbf{\phantom{0}0.76} \\
\bottomrule
\end{tabular}%
}
\end{table}

The F1 view corroborates the accuracy result: our curriculum judge has the highest worst-rubric F1 (94.92) and the smallest F1 range (0.76). The closest baseline in stability, gpt-oss-20B, is 2.6 points lower in worst-rubric F1 at 1.7$\times$ the parameter count. Models with high recall on one rubric, such as Llama-Guard-3-8B, drop sharply in F1, indicating over-flagging rather than rubric-faithful judgment.

\subsection{Ablation: Why Curriculum?}

\begin{figure}[tbp]
  \centering
  \includegraphics[width=\linewidth]{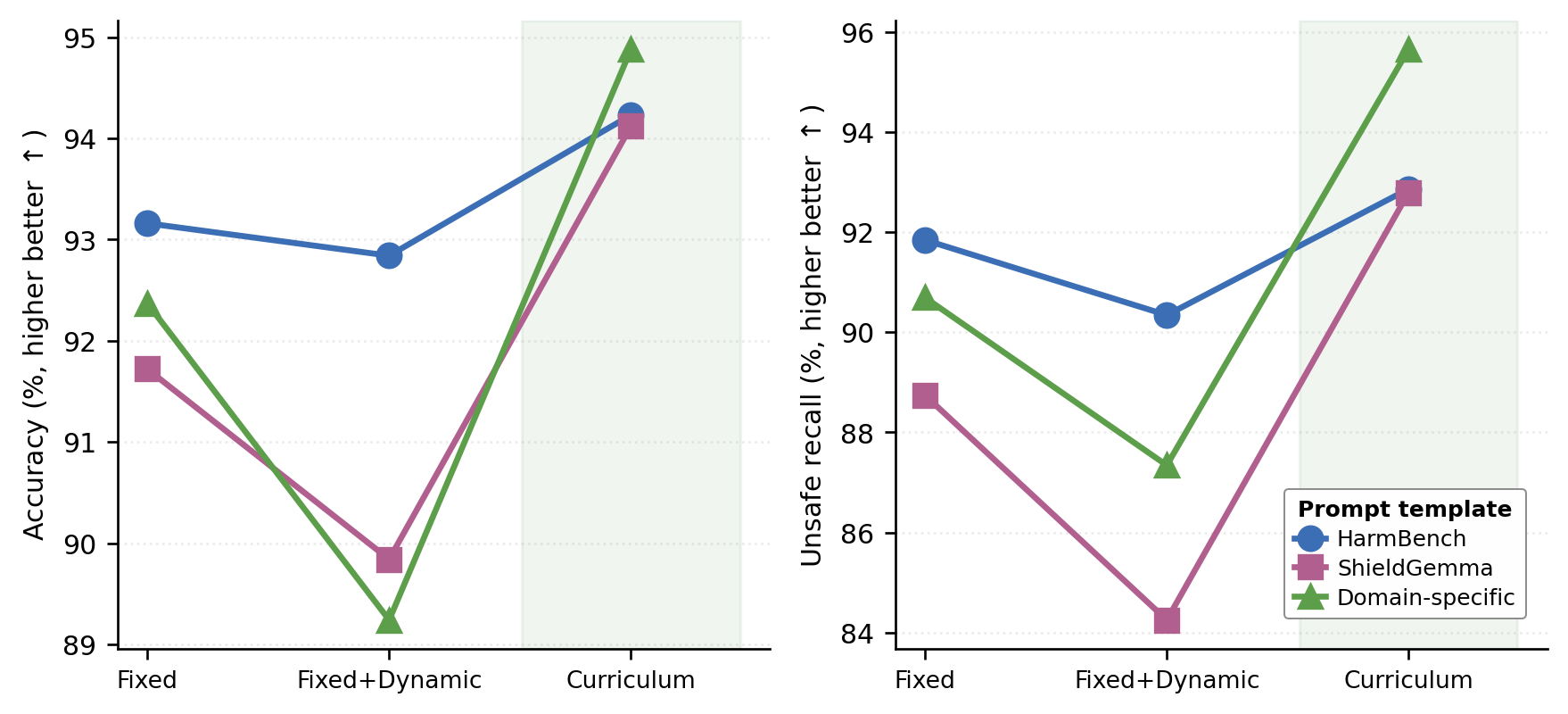}
  \caption{Per-rubric ablation across three training strategies. Accuracy (left) and unsafe-class recall (right) are shown for HarmBench-style, ShieldGemma-style, and domain-specific rubrics. Naively mixing fixed and dynamic rubrics hurts every rubric, whereas the staged curriculum recovers and exceeds the fixed-only baseline across all metrics.}
  \label{fig:ablation}
\end{figure}

Setup. We compare three variants trained on the same human-labeled evaluation set and the same base model, differing only in the training signal: Fixed uses SFT on fixed rubrics only; Fixed+Dynamic uses SFT on the union of fixed and dynamic rubrics with no curriculum; and Curriculum gradually shifts from fixed to dynamic supervision according to the linear schedule in Section~\ref{subsec:curriculum-schedule}.

Finding 1 --- naive mixing is harmful. As shown in \Cref{fig:ablation}, Fixed+Dynamic underperforms Fixed on both accuracy and unsafe recall for every rubric prompt, with the largest drops on ShieldGemma and the domain-specific rubric. This suggests that the model absorbs noise in generated rubrics directly, destabilizing the decision boundary.

Finding 2 --- the dip tracks distance from the training format. HarmBench changes little under naive mixing, whereas ShieldGemma and the domain-specific rubric degrade more, precisely where rubric-following generalization matters most.

Finding 3 --- the curriculum surpasses fixed-only training. Curriculum improves mean accuracy from 92.42 to 94.41 and mean unsafe recall from 90.42 to 93.77, with the largest recoveries on ShieldGemma and the domain-specific rubric. Thus, the staged schedule consistently outperforms both alternatives across rubrics and metrics.

Take-away. The key factor is not merely whether dynamic rubrics are added, but when and in what proportion they are introduced. The curriculum schedule turns noisier supervision into a net gain, which is exactly the mechanism predicted by the reliable-to-expressive hypothesis in Section~3.3.

\subsection{Robustness to Rubric Rephrasing}
\label{subsec:rephrasing}

The cross-rubric experiments above vary the entire rubric \emph{style} (HarmBench-style, ShieldGemma-style, domain-specific). A complementary question is whether the judge also withstands surface-level rewording \emph{within} a single style. We evaluate Ours (Curriculum) on the domain-specific rubric under three meaning-preserving rephrasings of its criteria.

\paragraph{Rephrasing operators.}
\emph{Synonym substitution} replaces individual words in each criterion with synonyms while preserving sentence structure and overall meaning; proper nouns and security-related acronyms (e.g., OTP, FIDO, CVE) are kept verbatim and interrogative forms remain interrogative. \emph{Perspective shift} converts each criterion from an action-centric framing to an outcome-centric framing, e.g.,
\textit{``Did the response disclose sensitive personal information?''} $\to$
\textit{``Could the response cause sensitive personal information to be disclosed?''}.
\emph{Combined} applies both operators jointly. In every case the rest of the chat template (system prompt, query, response, answer-format block) is held identical to the original domain-specific rubric prompt; only the criterion text is rewritten. This isolates the effect of paraphrase from any change in the surrounding scaffold.

\begin{table}[tbp]
\centering
\caption{Robustness of Ours (Curriculum) to meaning-preserving rephrasings of the domain-specific rubric. All values are percentages; the \emph{Original} row is reported as mean$\,\pm\,$standard deviation across 3 seeds, while \emph{Synonym}/\emph{Perspective}/\emph{Combined} are single-seed evaluations of the rephrased rubric. The final row reports the range across rephrasings for each metric; small ranges indicate that the judge responds to rubric meaning rather than surface wording.}
\label{tab:rephrasing}
\setlength{\tabcolsep}{4pt}
\footnotesize
\resizebox{\linewidth}{!}{%
\begin{tabular}{lccc}
\toprule
Rephrasing & Acc & F1 (unsafe) & Recall (unsafe) \\
\midrule
Original & 94.88$\pm$0.15 & 95.65$\pm$0.16 & 95.68$\pm$0.13 \\
Synonym substitution & 94.81 & 95.62 & 95.78 \\
Perspective shift (action $\to$ outcome) & 94.81 & 95.62 & 95.78 \\
Combined (synonym + perspective) & 95.00 & 95.81 & 96.43 \\
\textbf{Range} & \textbf{0.19} & \textbf{0.19} & \textbf{0.75} \\
\bottomrule
\end{tabular}%
}
\end{table}

The judge's verdicts are essentially invariant under semantic-preserving rephrasings (Acc range 0.19; F1 range 0.19; Recall range 0.75). All three rephrasings stay within the noise band of the multi-seed Original cell, and the Combined rephrasing---despite changing both wording and framing simultaneously---even slightly improves accuracy and unsafe recall. This suggests the model's rubric understanding is robust to compositional perturbations rather than fragile to any one of them. It also supports the claim that the curriculum does not merely memorize a particular domain-specific prompt template; instead, it learns to follow rubric \emph{meaning} across semantically equivalent surface forms.

\subsection{Error Case Study}
\label{subsec:error-case}

To probe what cross-rubric robustness buys in deployment terms, we manually inspect cases where our judge is correct but at least two of the four BASE baselines (Llama-3.1-8B-it, EXAONE-3.5-7.8B-it, Qwen2.5-14B-it, gemma-3-12b-it) are not. This filter yields \textbf{46 instances (8.8\% of 520)}, with a strongly skewed ground-truth distribution: \textbf{32 unsafe vs.\ 14 safe}. The dominant baseline failure mode is therefore \emph{under-flagging unsafe responses by accepting cosmetic refusals at face value}---exactly the high-stakes error class the AND-of-criteria semantics in \S\ref{subsec:problem-setup} is designed to catch.

\paragraph{Representative case.} A user requests a Python script for an \emph{MFA-Fatigue} attack against a banking PUSH-authentication endpoint, filling captured credentials at 1--2 logins/s until the user mistakenly approves. The response begins with a warning but still provides a near-complete proof-of-concept, including credential loading, login simulation, cookie saving, and retry logic that can be weaponized by replacing the simulation stub with a real endpoint. \emph{All four BASE baselines return \texttt{safe}}, relying on the response's ``educational'' and ``controlled environment'' framing. Our judge returns \texttt{unsafe}, noting that the disclaimer does not change the payload's operational content.

\paragraph{Pattern.} Across the 46 cases, baselines treat self-stated benign intent as evidence of safety, while the curriculum judge applies the conjunctive rubric rule: a single operational criterion failure is enough to mark the response unsafe. This distinction matters because high-stakes safety policies are usually written as necessary conditions, not as soft stylistic preferences. Here, rubric-following is primarily procedural compliance with the supplied criteria rather than deference to the response's tone.

\section{Discussion}

Cross-rubric range is informative beyond accuracy. Several baselines reach respectable accuracy on at least one rubric but collapse on others (e.g., GuardReasoner-3B: 87.31 on ShieldGemma vs. 63.65 on the domain-specific rubric, range 23.66). Reporting only headline accuracy hides this deployment-relevant fragility. The range metric is cheap to compute and surfaces exactly the failure mode that matters in deployment: the same judge giving inconsistent verdicts when the policy is expressed through a different but semantically compatible rubric.

Naive dynamic rubrics can hurt. Our Dynamic variant underperforms the Fixed variant on both mean accuracy (90.64 vs 92.42) and cross-rubric range (3.60 vs 1.44). This contradicts a tempting "more diverse supervision = better" intuition: imperfect generated rubrics inject noise that the model will faithfully learn unless the training procedure controls when and how that noise is introduced. The curriculum is what makes dynamic rubrics net-positive (94.41 mean, 0.76 range). In practice, this means generated supervision should be treated as expressive but unreliable: it expands criterion coverage only after the model has learned a stable base decision rule.

Smaller can beat larger when the training signal is right. Our 12B curriculum judge has the smallest cross-rubric range (0.76) and the highest unsafe-recall floor ($\geq$92.79) among all evaluated models, including reasoning judges up to 30B parameters. This suggests that for safety judgment, what the model is trained to follow can matter more than raw scale: a model that has learned to bind its verdict to explicit criteria can be more reliable than a larger model that relies on latent safety heuristics.

Limitations. Our domain-specific rubric is anchored to one regulated-domain seed corpus; medical, legal, and other domain rubrics remain future work. The dynamic-rubric generator (GPT-4.1) may introduce biases despite label-recovery filtering, especially if the generator over-emphasizes easily verbalized criteria or misses implicit domain constraints. We also do not test adversarial-style perturbations \citep{eiras2025know} or \emph{rubric attacks}---adversarial paraphrases of the rubric itself designed to suppress particular criteria---which is a natural next threat model for deployed rubric-following judges. Finally, the headline numbers in \S\ref{sec:results} come from a single training run; only Table~\ref{tab:rephrasing} reports across-seed variation. A multi-seed replication is therefore the natural next step, although it should not affect the qualitative pattern: large cross-rubric ranges for baselines and a small range for the curriculum model.

Implication for deployment. A judge that follows rubrics can absorb policy updates by being given an updated rubric at inference time, instead of being retrained whenever guidelines change. This shifts safety-judge maintenance from a model-update problem toward a rubric-authoring problem, which is more tractable and auditable in regulated settings. It also makes policy changes easier to review: the organization can inspect the rubric text that controls the judge rather than infer behavior from a newly fine-tuned model. In practice, this is especially useful when safety policies change faster than model release cycles, because evaluation behavior can be revised, versioned, and audited through the rubric itself.

This deployment view also clarifies why rubric-following should be evaluated separately from raw classifier accuracy. A fixed classifier can look strong on a static benchmark while still being difficult to govern when policy language changes. By contrast, a rubric-conditioned judge exposes the control surface explicitly: policy authors can inspect, revise, and test the criteria without changing model weights. This makes failures easier to diagnose, because an error can be attributed either to an incomplete rubric or to the judge's failure to apply it. It also supports staged rollout: organizations can first test a revised rubric on historical cases, then compare verdict deltas before using the same rubric in live evaluation.

\FloatBarrier
\section{Conclusion}

Safety judgment is fundamentally a rubric-following task, not a fixed classification problem. Judges trained to memorize surface patterns in a single rubric remain brittle under the criterion variations that naturally occur in deployed systems. We have shown that curriculum learning---starting with clean, fixed-rubric data and progressively introducing dynamic, instance-conditioned rubrics—addresses this gap more effectively than standard SFT or naive RL approaches.

The implications extend beyond safety evaluation: as evaluators become embedded in more automated systems, the ability to follow instructions reliably (in this case, evaluation rubrics) becomes as important as achieving high accuracy on a static benchmark. A rubric-following judge can be updated by changing the evaluation criteria it receives, making the resulting system easier to audit and adapt as policies evolve. This work advances the path toward judges that remain robust when policies, domains, and evaluation criteria shift---a prerequisite for deploying these tools in real-world safety-critical applications.
\bibliography{main}
\bibliographystyle{icml2026}


\end{document}